\documentclass{article}

\usepackage[final, nonatbib]{neurips_2023}

\usepackage[utf8]{inputenc} %
\usepackage[T1]{fontenc}    %
\usepackage[table, dvipsnames]{xcolor}
\definecolor{citecolor}{HTML}{021971}
\usepackage[pagebackref=true,breaklinks=true,letterpaper=true,colorlinks,
citecolor=citecolor,bookmarks=false]{hyperref}

\usepackage{url}            %
\usepackage{booktabs}       %
\usepackage{amsfonts}       %
\usepackage{nicefrac}       %
\usepackage{microtype}      %

\usepackage{times}
\usepackage{epsfig}
\usepackage{graphicx}
\usepackage{amsmath}
\usepackage{amssymb}
\usepackage{subfiles}
\usepackage{makecell} 
\usepackage{wrapfig} 
\usepackage{bm}
\usepackage{subfigure}

\usepackage{physics}

\usepackage{array}
\usepackage{multirow}

\newlength\savewidth

\usepackage{xspace}

\newcommand{\dataset}[0]{\textbf{\texttt{Wusi}}\xspace}

\makeatletter
\DeclareRobustCommand\onedot{\futurelet\@let@token\@onedot}
\def\@onedot{\ifx\@let@token.\else.\null\fi\xspace}

\def\eg{\emph{e.g}\onedot} 
\def\ie{\emph{i.e}\onedot} 
 
\def\etc{\emph{etc}\onedot} 
 
\def\etal{\emph{et al}\onedot}
\makeatother

\newcolumntype{S}{>{\centering\arraybackslash}m{0.9cm}}
\newcolumntype{M}{>{\centering\arraybackslash}m{1.2cm}}
\newcolumntype{L}{>{\centering\arraybackslash}m{1.4cm}}
\definecolor{mygray}{gray}{.95}
\definecolor{mylightergray}{gray}{.99}
\definecolor{mygreen}{RGB}{10, 179, 33}
\usepackage{caption} 
\makeatletter
\newcommand{\thickhline}{%
    \noalign {\ifnum 0=`}\fi \hrule height 1pt
    \futurelet \reserved@a \@xhline
}
\newcolumntype{"}{@{\vrule width 1pt}}

\definecolor{mygray}{gray}{.95}
\definecolor{mylightergray}{gray}{.99}
\definecolor{mygreen}{RGB}{10, 179, 33}

\usepackage{pifont}

\usepackage[capitalize]{cleveref}
\crefname{section}{Sec.}{Secs.}
\Crefname{section}{Section}{Sections}
\Crefname{table}{Table}{Tables}
\crefname{table}{Tab.}{Tabs.}

\title{Social Motion Prediction with Cognitive Hierarchies}

\author{%
  Wentao Zhu$^{\,1}\thanks{Equal Contribution.}\footnotemark[1]$ 
  \quad
    Jason Qin$^{\,1}\footnotemark[1]$ 
  \quad
    Yuke Lou$^{\,1}$
  \quad 
    Hang Ye$^{\,1}$
  \\
    \textbf{Xiaoxuan Ma}$^{\,1}$
  \quad
    \textbf{Hai Ci}$^{\,1}$
  \quad
    \textbf{Yizhou Wang}$^{\,1,\,2}$ \\
    \and
    $^1$ Center on Frontiers of Computing Studies, School of Computer Science, Peking University\\
    $^2$ Institute for Artificial Intelligence, Peking University\\
}

\begin{document}

\maketitle

\begin{abstract}

Humans exhibit a remarkable capacity for anticipating the actions of others and planning their own actions accordingly. In this study, we strive to replicate this ability by addressing the social motion prediction problem. We introduce a new benchmark, a novel formulation, and a cognition-inspired framework. We present \dataset, a 3D multi-person motion dataset under the context of team sports, which features intense and strategic human interactions and diverse pose distributions. By reformulating the problem from a multi-agent reinforcement learning perspective, we incorporate behavioral cloning and generative adversarial imitation learning to boost learning efficiency and generalization. Furthermore, we take into account the cognitive aspects of the human social action planning process and develop a cognitive hierarchy framework to predict strategic human social interactions. We conduct comprehensive experiments to validate the effectiveness of our proposed dataset and approach. Code and data are available at \url{https://walter0807.github.io/Social-CH/}.

\end{abstract}

\section{Introduction}
\label{sec:intro}

Human beings are inherently social creatures. Concretely, individuals unconsciously anticipate the actions of others and make informed decisions about their own behaviors in social contexts~\cite{chartrand1999chameleon, frith2006neural, sebanz2006joint}, which enables individuals to cooperate and compete with others in a variety of scenarios, from pedestrian traffic~\cite{helbing2001self, moussaid2010walking} to team sports~\cite{vilar2012role, ric2016soft}. Notably, task experts demonstrate exceptional skill in predicting others' movements in advance~\cite{loffing2017anticipation, williams2002anticipation}.

To better understand and replicate this ability, the research community has proposed the task of future prediction for multiple interacting agents given their historical observations. The majority of prior work in this area focuses on modeling and predicting agent interactions at the trajectory level~\cite{mohamed2020social,mangalam2020not,sadeghian2019sophie,amirian2019social,kosaraju2019social,sun2020recursive,yu2020spatio,sun2021three,implicit,socialvae2022} and has demonstrated promising results in applications such as autonomous driving~\cite{luo2018porca,raksincharoensak2016motion,deo2018convolutional,li2019grip,cui2019multimodal,huang2023gameformer}. However, trajectory-based approaches can only reflect coarse-grained interactions (\eg, collision avoidance, social distancing) and fail to capture the rich and fine-grained human actions. To this end, some studies have investigated the \emph{multi-person motion prediction} problem, which aims to forecast both trajectories and poses for a group of people~\cite{scmpf,tripod,futuremotion,mrt,xia,vendrow-somoformer,peng-somoformer,xu2023stochastic}.
Despite recent advancements, there remain several critical challenges in this field. Firstly, existing multi-person motion datasets are primarily designed for human pose estimation tasks~\cite{3dpw,mocap,panoptic}, and consequently, do not place particular emphasis on human interactions. Individuals in these datasets tend to move casually and interact with others at random, making future predictions both difficult and less meaningful. Secondly, the majority of prior methods concentrate on developing neural network architectures for end-to-end supervised training while overlooking the cognitive aspects of human social action planning. These two challenges are closely related and necessitate a comprehensive solution.

\begin{figure*}[t]
  \centering
  \includegraphics[width=\linewidth]{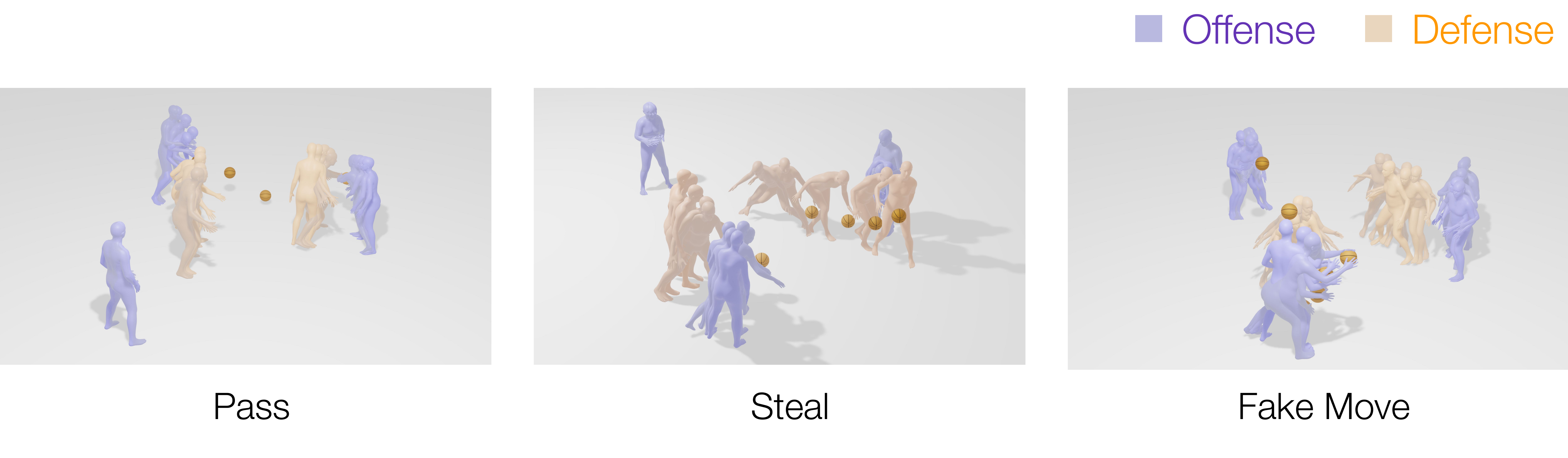}
  \caption{Example sequences from \dataset dataset. Three players in purple form the offensive team, and two players in orange form the defensive team. Left: a successful pass for the offensive team. Middle: a successful steal for the defensive team. Right: a successful pass following a fake pass. 
}
  \label{fig:dataset}
  \vspace{-0.4cm}
\end{figure*}

In this work, we propose novel perspectives on this problem to address the aforementioned limitations. We begin by constructing a large-scale multi-person 3D motion dataset featuring intense and strategic interactions among participants. To achieve this, we turn to team sports, which offer several inherent advantages: 1) Well-defined game rules and global rewards implicitly constrain and guide individual actions. 2) Participants develop intricate interaction strategies based on their roles, such as planning cooperative actions with teammates while acting adversarially towards opponents. Skilled players even employ sophisticated techniques like deception, \etc. 3) Human motions exhibit greater dynamism in terms of pose diversity and motion intensity, making motion prediction more challenging than in previous datasets.

Additionally, we present a new formulation of the multi-person motion prediction task as a multi-agent reinforcement learning (MARL) problem. Specifically, we model the task using imitation learning, where the objective is to learn a policy from expert demonstrations. We employ behavioral cloning (BC) \cite{Bain95, Ross11} to imitate expert behaviors from the dataset. To improve learning efficiency and generalization, we utilize generative adversarial imitation learning (GAIL) \cite{Ho16}, aiming to render agents' policy indistinguishable from experts' policy. Furthermore, we propose a framework for human social interactions based on the cognitive hierarchy theory~\cite{camerer2004cognitive}. In particular, we assume that people base their decisions on their predictions regarding the likely actions of others, while others engage in similar decision-making processes from their perspectives. By considering the reasoning steps recursively, we posit that a level $k$ agent takes actions based on the level $k-1$ agent actions, where $k$ represents the depth of strategic thought. Building on this insight, we develop a computation model that can be elegantly integrated with the MARL formulation.

We summarize our contributions as follows: 1) We present \dataset, the first large-scale multi-human 3D motion dataset featuring intense and strategic interactions. We demonstrate that our dataset has greater motion diversity than existing datasets and poses a more significant challenge for the social motion prediction problem. 2) We propose a novel MARL formulation for the problem and develop an imitation learning baseline that combines behavioral cloning and generative adversarial imitation learning. 3) We introduce a cognitive hierarchy framework to model the strategic and game-theoretic human social interactions. Our approach outperforms the state-of-the-art methods in challenging long-term social motion predictions.

\section{Related work}
\label{sec:related-work}

\subsection{Multi-agent trajectory prediction}
The multi-agent trajectory prediction problem has been extensively studied, especially for interacting traffic participants, \eg, pedestrians and vehicles. There exist multiple industrial benchmarks including Argoverse~\cite{chang2019argoverse}, nuScenes~\cite{caesar2020nuscenes}, Waymo Open Motion Dataset~\cite{ettinger2021large}. Meanwhile, various generation methods have been explored to model the interaction among different agents and predict their future trajectories~\cite{mohamed2020social,mangalam2020not,sadeghian2019sophie,amirian2019social,kosaraju2019social,sun2020recursive,yu2020spatio,sun2021three,implicit,socialvae2022}. 
In this work, we aim to predict intense and strategic interactions for a group of humans with fine-grained body motions.

\subsection{Multi-person motion dataset}

The existing multi-human motion datasets can be categorized into 2D and 3D. The PoseTrack dataset~\cite{posetrack,posetrack2017} offers video sequences with manually annotated 2D keypoints. 
However, 2D keypoints fail to accurately represent the proximity of individuals in the real world, and manual annotation is difficult to scale. Meanwhile, 3D multi-person motion datasets can be effectively constructed with motion capture (mocap) systems, including marker-based solutions and markerless ones. 3DPW~\cite{3dpw} uses the inertial measurement units (IMUs) to obtain high-quality motion reconstructions. UMPM~\cite{van2011umpm} employs a set of reflective markers to identify the joint positions.
However, wearable sensors may struggle to capture complex movements and become expensive as the number of individuals increases. Other datasets, \eg, CMU-Mocap~\cite{mocap}, Panoptic~\cite{panoptic}, MuPoTs-3D~\cite{mupots}, ExPI~\cite{xia}, employ markerless mocap systems with multi-view cameras to address the occlusion problem and obtain 3D human motion through triangulation. 
In this work, we construct a multi-person motion dataset featuring high motion diversity and intense social interactions using multi-view cameras. We provide a comparison with previous datasets in Section~\ref{data_analysis}.

\subsection{Multi-person motion prediction}

Human motion prediction for a single person has been extensively studied ~\cite{ltd, hri, chen2023humanmac} in previous research. Recently, however, there has been a growing interest in multi-person motion prediction, which further involves the prediction of nuanced social interactions.
Joo~\etal~\cite{joo2019towards} propose to predict human motion conditioned on other individuals' kinesic signals in a triadic haggling scenario.
Adeli~\etal~\cite{scmpf} introduce the 2D social motion forecasting (SoMoF) benchmark, which aims to predict multi-person trajectory and pose on the PoseTrack dataset \cite{posetrack}. They also propose a baseline model using a shared GRU encoder and a pooling layer to incorporate social clues.
TRiPOD~\cite{tripod} employs attention graphs to characterize the spatiotemporal social interactions. It also considers the joint occlusion and body invisibility issues arising from 2D observations.
Futuremotion~\cite{futuremotion} provides a simple baseline using no social context, yet achieving competitive performance on SoMoF.
As 2D multi-person motion prediction faces challenges such as data scarcity, depth ambiguity, and occlusions, more recent research has shifted towards exploring the multi-person motion prediction problem in 3D. Wang~\etal~\cite{mrt} propose a multi-range Transformer to separately encode the local and global motion history. 
Guo~\etal~\cite{xia} design a cross-interaction-attention (XIA) module to model close interactions between pairs in duo dance scenarios. 
Vendrow~\etal~\cite{vendrow-somoformer} introduce joint-aware attention and joint-wise query to predict the entire future sequence without recurrence.
Peng~\etal~\cite{peng-somoformer} present a social-aware motion attention mechanism that models both inter- and intra-individual motion relations. 
DuMMF~\cite{xu2023stochastic} frames the problem as a dual-level generative task and instantiates it with various generative models.
In contrast to previous work, we adopt a MARL perspective and model the strategic social interaction process using cognitive hierarchies.

\subsection{Cognitive hierarchy theory}
Cognitive hierarchy theory (CHT) is a model in behavioral economics and game theory that aims to describe human decision processes in strategic games. Several foundational works introduce the concept of CHT~\cite{camerer2004cognitive}, and provide experimental evidence~\cite{nagel1995unraveling,costa2006cognition,haruvy1999evidence,stahl1994experimental,stahl1995players} supporting the existence of cognitive hierarchies in human decision-making processes. 
Taking it a step further, Wright~\etal~\cite{wright2010beyond} develop a Poisson cognitive hierarchy model to predict human behavior in normal-form games. Li~\etal ~\cite{li2022role} summarize how the information structure guided by the cognitive hierarchy supports belief generation and policy generation in game-theoretic multi-agent learning. In this work, we propose a cognitive hierarchy framework in conjunction with MARL, and demonstrate its effectiveness in a real-world social motion prediction scenario.

\section{Dataset}    
\label{sec:dataset}

\subsection{Overview}
We present a multi-person 3D motion dataset with a special focus on strategic interactions called \dataset (Wusi Basketball Training Dataset). In the following, we first introduce the contents of the dataset, then outline the data collection pipeline, and finally provide statistical analysis and comparison with regard to existing datasets.

Our dataset captures the no-dribble-3-on-2 basketball drills performed by a team of professional basketball athletes. In each drill, three offensive players possess the ball, while another two players are on defense. The offensive team aims to accomplish as many successful passes as possible within the given time, while the defensive team strives to get steals, deflections, and slow down the offense. Since dribbling is prohibited, the drill requires the offensive players to make better passing decisions and foresee the defense before they pass; conversely, the defensive players need to anticipate the passing directions in order to steal the ball. 
Figure~\ref{fig:dataset} showcases the diverse and dynamic human interactions present in our dataset. On the left, the offensive player successfully passes the ball to a teammate. In the middle, the defensive player accurately anticipates the passing trajectory and successfully steals the ball. On the right, the offensive player pretends to pass the ball to the right, then swiftly executes a pass to the left.

\subsection{Data collection}

Our motion capture system consists of 11 synchronized and calibrated wide-baseline cameras. Our setup ensures that the players are well-surrounded by the cameras from elevated shooting angles, enabling each body joint to be covered by at least four different camera views. We utilize a markerless multi-person 3D pose estimation algorithm~\cite{fvox} with JDETracker~\cite{jde}. Subsequently, we apply a post-processing pipeline that includes automatic failure detection and temporal filtering~\cite{oneeuro}. A final manual check is conducted to guarantee the integrity of the data sequences.

\begin{table}[t]
\centering
\caption{Dataset comparison. We compare our dataset with existing multi-person motion datasets employed by previous works on the multi-person motion prediction task. $^\dagger$ denotes multi-person subset as utilized in previous works ~\cite{scmpf, mrt}.
}
\label{dataset-comparison}
\resizebox{0.75\linewidth}{!}{%
\scriptsize
\setlength{\tabcolsep}{3pt}
\begin{tabular}{@{}lcccccc@{}}
\toprule
Dataset & 2D/3D & Frames & Duration (min) & No. of people & Interaction \\ \midrule
PoseTrack~\cite{posetrack}$^\dagger$ & 2D & 8K & 5.5 & Multiple & Weak \\
CMU-Mocap~\cite{mocap}$^\dagger$ & 3D & 34K & 4.9 & 2 & Weak \\
3DPW~\cite{3dpw}$^\dagger$ & 3D & 5K & 6.1 & 2 & Weak \\
MuPoTs-3D~\cite{mupots} & 3D & 8K & 4.4 & 2-3 & Weak \\
ExPI~\cite{xia} & 3D & 30K & 20.0 & 2 & Cooperative \\
\dataset & 3D & 60K & 40.3 & 5 & Strategic \\ \bottomrule
\end{tabular}
}
\end{table}

\begin{table}[h]
\centering
\caption{Comparison of pose diversity based on different thresholds.}
\label{tab:comparison}
\begin{tabular}{lcc}
\toprule
Threshold     & 50mm  & 100mm \\
\midrule
Human3.6M~\cite{Ionescu14} & $24\%$  & $12\%$  \\
ExPI~\cite{xia}     & $52\%$  & $23\%$  \\
CMU-Mocap~\cite{mocap} & $20\%$  & $9\%$   \\
MuPoTs-3D~\cite{mupots} & $37\%$  & $19\%$  \\
\dataset          & $\mathbf{53\%}$ & $\mathbf{27\%}$ \\
\bottomrule
\end{tabular}
\label{tab:diversity}
\end{table}

\subsection{Data analysis}
\label{data_analysis}

In this section, we examine the statistics of \dataset dataset and compare it with existing datasets for multi-person motion prediction. As outlined in Table~\ref{dataset-comparison}, current datasets exhibit two primary limitations for the multi-person motion prediction task. Firstly, high-quality 3D motion data is limited in terms of scale (duration and the number of people). Consequently, prior works~\cite{mrt, peng-somoformer, vendrow-somoformer, xu2023stochastic} resort to randomly mixing the motion sequences~\cite{mocap,3dpw,mupots} in order to generate more multi-person motion sequences for training. However, such practice inevitably causes unnatural human interactions. Secondly, the interaction strength in existing datasets is restricted, mainly featuring simple actions such as walking together~\cite{posetrack, mocap, 3dpw, mupots}. More recently, the ExPI dataset~\cite{xia} introduces cooperative 3D motions of two Lindy-hop dancers. In contrast, our dataset offers 2 to 9 times the video duration of existing datasets and includes strategic global interactions among 5 individuals.

\begin{figure*}[t]
  \centering
  \includegraphics[width=\linewidth]{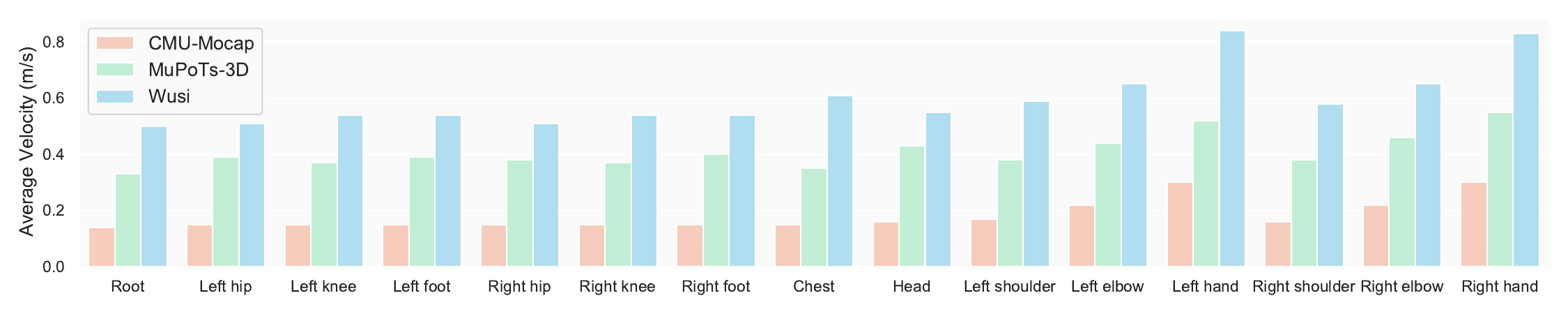}
  \caption{Comparison of motion intensity.
  }
  \label{fig:intensity}
\end{figure*}

Additionally, we provide a quantitative comparison with existing datasets, emphasizing two critical aspects: \emph{pose diversity} and \emph{motion intensity}. 
Firstly, we examine the pose diversity within the datasets by calculating the ratio of unique poses to the total number of poses following ~\cite{Ionescu14, xia}. Table ~\ref{tab:diversity} shows that our dataset surpasses the previous datasets in terms of overall pose diversity.
We further analyze the motion intensity by computing the average velocity for all the body joints. 
As illustrated in Table ~\ref{fig:intensity}, our dataset exhibits more dynamic movements across all body joints.
We suppose that these characteristics make \dataset not only challenging for motion prediction, but also potentially useful for other related tasks, \eg motion generation~\cite{zhu2023human}.

\section{Method}
\label{sec:methods}

\subsection{Formulation}

The multi-person motion prediction problem can be formulated as follows: given the motion histories $\{x_t^p\}_{1 \leq t \leq T, 1 \leq p \leq P}$ of length $T$ from $P$ human subjects, predict their future motion $\{x_t^p\}_{T \leq t \leq T+T', 1 \leq p \leq P}$ of length $T'$, where each $x_t^p$ represents a 3D human pose.

We model the problem using a Markov Decision Process (MDP)~\cite{Sutton17} by dividing the prediction sequence into $L$ steps with even step length $m=\frac{T'}{L}$ following ~\cite{wang2019imitation}, and further extend the formulation to multiple agents. We use bold symbols to represent the collective variables of all agents following the convention in MARL.
At each step $i$, The state $\vb*{s}_i$ is defined as the aggregated motion history $\{x_t^i\}_{1 \leq t \leq T + (i-1) \times m, 1 \leq p \leq P}$ for all the agents. The action for the $p$-th agent $a^p$ is defined as a sequence of velocities $\{v_t^p\}_{T + (i-1) \times m \leq t \leq T + i \times m - 1}$, where $v_t^p=x_{t+1}^p-x_t^p$. 
Therefore, the joint action of all $P$ agents $\vb*{a}_i$ deterministically transition the MDP into the new state $\vb*{s}_{i+1}$ according to: 
\begin{equation}
\small
    x^p_{t^*} = x_{T + (i-1) \times M}^p + \sum_{t=T + (i-1) \times M}^{t^*} v_t^p ,\quad T + (i-1) \times m +  \leq t^* \leq T + i \times m, 1 \leq p \leq P.
\end{equation}

The goal of this objective is to learn a policy $\pi^p$ for each agent $p$ that maps each possible state $\vb*{s}$ to agent action $a^p$ by $a^p=\pi^p(\vb*{s})$. In the real world, people optimize their $\pi^p$ to maximize an implicit team reward $r$. Since we have no access to $r$ or the environment, we aim to learn $\pi^p$ from expert demonstrations $\mathcal{D}$ via imitation learning.

\begin{figure*}[t]
  \centering
  \includegraphics[width=\linewidth]{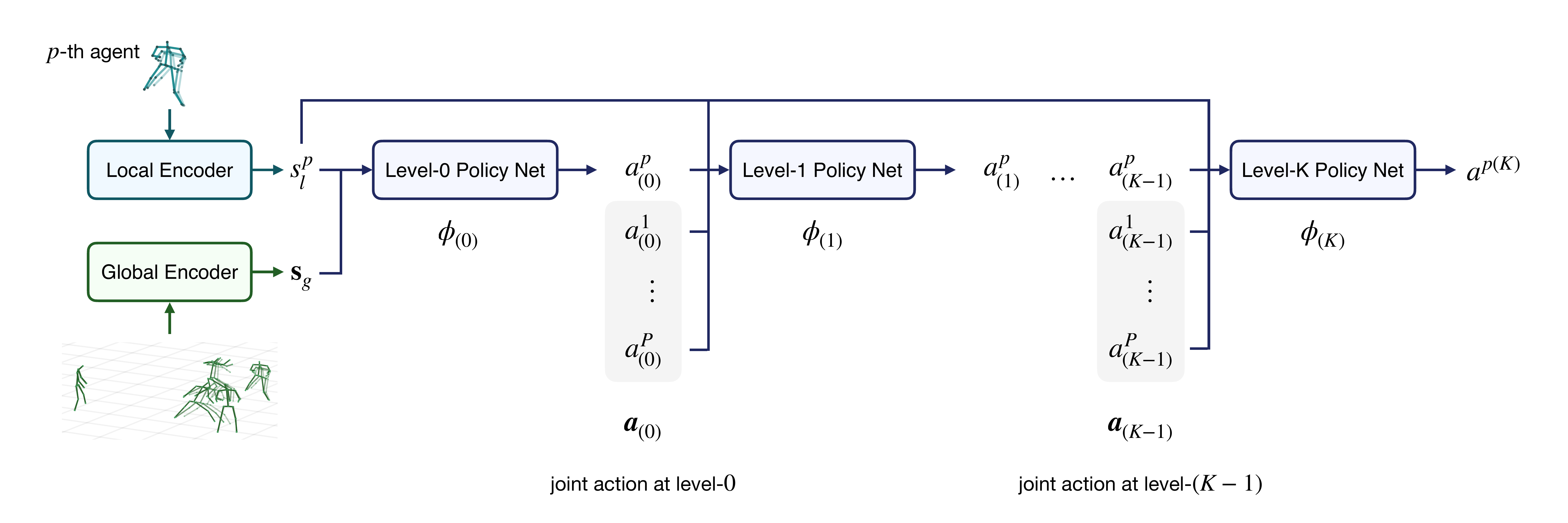}
  \caption{Framework overview. For the $p$-th agent, two state encoders first extract its local and global state features $s^p_l$ and $\vb*{s}_g$, from which the level-$0$ policy network produces an initial action $a^{p}_{(0)}$. The level-$k$ policy network produces action $a^{p}_{(k)}$ based on $s^p_l$ and the joint actions of the previous level $\vb*{a}_{(k-1)}$ ($k \geq 1$). 
  }
  \label{fig:arch}
  \vspace{-0.4cm}
\end{figure*}

We parameterize $\pi^p$ with a set of parameters $\theta$. In practice, we follow~\cite{mrt} to employ a global-range Transformer $E_g$ to encode global state feature $\vb*{s}_g=E_g(\vb*{s})$, and a local-range Transformer $E_l$ to encode local state feature $s^p_l=E_l(\vb*{s})$ for each agent $p$. See Figure \ref{fig:arch} for an overview. This approach helps to extract multi-level state features and enables all agents to share policy network parameters.

\subsection{Behavioral Cloning}
A straightforward method in imitation learning is Behavioral Cloning (BC)~\cite{Bain95, Ross11}, \ie, using expert demonstrations to minimize the difference between the actions produced by the policy and those taken by the experts:

\begin{equation}
\small
    \theta^*=\operatorname{argmin}_\theta \sum_{\mathcal{D}} \sum^{P}_{p=1} \ell\left(\pi^p_\theta\left(\vb*{s}\right), \tilde{\pi}^p\left(\vb*{s}\right) \right),
\end{equation}

where $s$ is sampled from expert demonstrations $\mathcal{D}$, $\pi^p_\theta$ and $\tilde{\pi}^p$ are policies of the model and the expert for agent $p$, $\ell$ is a distance function defined in the action space. While BC enjoys the benefit of being both computation-efficient and sample-efficient~\cite{sammut2010behavioral}, it also faces several drawbacks. For instance, policies tend to be overfitted with respect to the state distribution encountered by the experts, leading to suboptimal generalization ability~\cite{wang2019imitation, jena2021augmenting}. To cope with this challenge, we further introduce GAIL and cognitive hierachy reasoning.

\subsection{Generative Adversarial Imitation Learning}

Generative Adversarial Imitation Learning (GAIL)~\cite{Ho16} is a model-free imitation learning approach.
It leverages the adversarial training framework of Generative Adversarial Networks (GAN)~\cite{Goodfellow14} and the concept of inferring reward functions from Inverse Reinforcement Learning (IRL)~\cite{Ng00}. 
Direct estimation of reward signals from expert demonstrations can be very difficult; therefore, GAIL transforms the IRL problem into its equivalent dual problem of occupancy measure matching. Specifically, the policy network is regularized to match its state-action pair distribution to that of the experts' policy through adversarial training. 
We implement GAIL using a global discriminator $D$ parameterized by $\omega$, which differentiates the state-action pairs produced by agents' joint policy $\boldsymbol{\pi}$ and experts' joint policy $\tilde{\boldsymbol{\pi}}$.
The optimization objective can be formulated as:

\begin{equation}
\small
    \min _\theta \max _\omega \mathbb{E}_{\boldsymbol{\pi}_\theta}\left[ \log D_{\omega}\left(\vb*{s}, \boldsymbol{\pi}_\theta\left(\vb*{s}\right) \right)\right]+\mathbb{E}_{\tilde{\boldsymbol{\pi}}}\left[ \log \left(1-D_{\omega}\left(\tilde{\vb*{s}}, \tilde{\boldsymbol{\pi}}\left(\tilde{\vb*{s}}\right) \right)\right)\right],
\end{equation}

where $\vb*{s}$ and $\tilde{\vb*{s}}$ are sampled from expert demonstrations $\mathcal{D}$ independently.

\subsection{Cognitive hierarchies}

Furthermore, we notice that in real-world situations, individuals would predict others' behaviors by forming beliefs about others' policies and then acting accordingly to maximize their own payoffs~\cite{fudenberg1998theory, camerer2011behavioral}. This can be formulated as a hierarchical, game-theoretic decision-making process, in which lower-level agents adopt straightforward strategies, while higher-level players anticipate the strategies of lower-level agents and respond accordingly.
We explicitly model this recursive reasoning process to learn more interpretable and robust agent policies. We adopt a specific type of cognitive hierarchy model~\cite{camerer2004cognitive} called \emph{level-k thinking}~\cite{stahl1994experimental, nagel1995unraveling, crawford2007level} to represent this process. Specifically, agents at each level (except for the lowest one) assume that others are reasoning at the previous level.
As shown in Figure ~\ref{fig:arch}, a straightforward policy is to take actions based on local and global state features $s^p_l$ and $\vb*{s}_g$, which we denote as level-$0$ policy $\pi^{p}_{(0)}$. Level-$0$ actions are thus defined as:

\begin{equation}
\small
    a^{p}_{(0)} = \pi^{p}_{(0)}(\vb*{s}) = \phi_{(0)}(s^p_l, \vb*{s}_g),
\end{equation}
where $\phi_{(0)}$ is the level-$0$ policy network. Then for a level-$k$ agent ($k\geq 1$), it takes actions based on the joint agent actions $\vb*{a}_{(k-1)}$ at the previous level and its local state features $s^p_l$:

\begin{equation}
\small
    a^{p}_{(k)} = \pi^{p}_{(k)}(\vb*{s}) = \phi_{(k)}(s^p_l, \vb*{a}_{(k-1)}), \quad 1\leq k \leq K
\end{equation}

where $\phi_{(k)}$ is the level-$k$ policy network, $K$ is the maximum strategic depth.

\subsection{Training objectives}
Finally, we introduce our training objectives which effectively integrate the aforementioned components. For level-$k$ ($1\leq k \leq K$) agents, we apply GAIL to regularize the distance between their joint policy and the experts' policy:

\begin{equation}
\small
    \mathcal{L}_{\text{GAIL}} =  \sum^{K}_{k=1} \mathbb{E}_{\boldsymbol{\pi}_{(k)}}\left[  \log D\left(\vb*{s}, \boldsymbol{\pi}_{(k)} \left(\vb*{s}\right) \right)\right]+\mathbb{E}_{\tilde{\boldsymbol{\pi}}}\left[ \log \left(1-D\left(\tilde{\vb*{s}}, \tilde{\boldsymbol{\pi}}\left(\tilde{\vb*{s}}\right) \right)\right)\right].
\label{l_gail}
\end{equation}

In addition, we employ BC on level-$K$ agent policies:

\begin{equation}
\small
    \mathcal{L}_{\text{BC}} =  \sum_{\mathcal{D}} \sum^{P}_{p=1} \ell\left(\pi^{p}_{(K)}\left(\vb*{s}\right), \tilde{\pi}^p\left(\vb*{s}\right) \right).
\end{equation}

We train the discriminator $D$ and the policy networks $\boldsymbol{\phi}$ alternatively, where $D$ aims to maximize $\mathcal{L}_{\text{GAIL}}$, and $\boldsymbol{\phi}$ is optimized to minimize a linear combination of $\mathcal{L}_{\text{BC}}$ and $\mathcal{L}_{\text{GAIL}}$ using a policy gradient algorithm following~\cite{jena2021augmenting}:

\begin{equation}
\small
    \mathcal{L}_{\boldsymbol{\phi}} = \mathcal{L}_{\text{BC}} + \lambda \mathcal{L}_{\text{GAIL}},
\end{equation}

where $\lambda$ is a constant for balancing the loss terms.

\section{Experiments}
\label{sec:experiments}

\subsection{Setup}
\noindent\textbf{Implementation details.}
We implement the presented framework to train and test on the proposed \dataset dataset. We employ Transformer encoder~\cite{vaswani2017attention} for both the local and global state encoders, as well as Transformer decoders for the policy networks. Each Transformer consists of $3$ layers with $8$ attention heads. We share parameters for policy networks $\phi_{(1)} \dots \phi_{(K)}$. We set the strategic reasoning depth $K=3$ unless otherwise stated. For all the methods, we provide $1$s motion history and predict future $1$s motion. For additional experimental details, please refer to the appendix.

\noindent\textbf{Evaluation metrics.}
We compute the Mean Per Joint Position Error (MPJPE) between the predicted future motion and ground-truth. In order to disentangle the results of trajectory prediction and pose prediction, we further calculate the mean root position error and mean local pose error (MPJPE after root alignment). We report all the errors in millimeters.

\subsection{Evaluation and comparison}

In this section, we evaluate the proposed dataset and method, comparing with the prior works. We conduct experiments with the following baseline methods including a naive baseline and multiple state-of-the-art approaches:
1) \emph{Frozen} is a naive baseline that simply replicates the last frame of the input motion. 
2) \emph{Social-STGCNN}~\cite{mohamed2020social} is a multi-agent trajectory prediction method using spatio-temporal graph convolutional networks. For trajectory prediction methods, we train and test using the root (global) trajectories and replicate the last frame for the root-relative (local) pose.
3) \emph{SocialVAE}~\cite{socialvae2022} is a multi-agent trajectory prediction method using timewise variational autoencoder. 
4) \emph{History Repeats Itself (HRI)}~\cite{hri} is a single-person motion prediction method that allows absolute coordinate inputs.
5) \emph{Multi-Range Transformers (MRT)}~\cite{mrt} predicts multi-person motion utilizing a local-range encoder and a global-range encoder.
6) \emph{Social Motion Transformer (SoMoFormer)}~\cite{vendrow-somoformer} employs the joints of all people as queries and predicts their future motion in parallel.

\begin{figure*}[t]
  \centering
  \includegraphics[width=\linewidth]{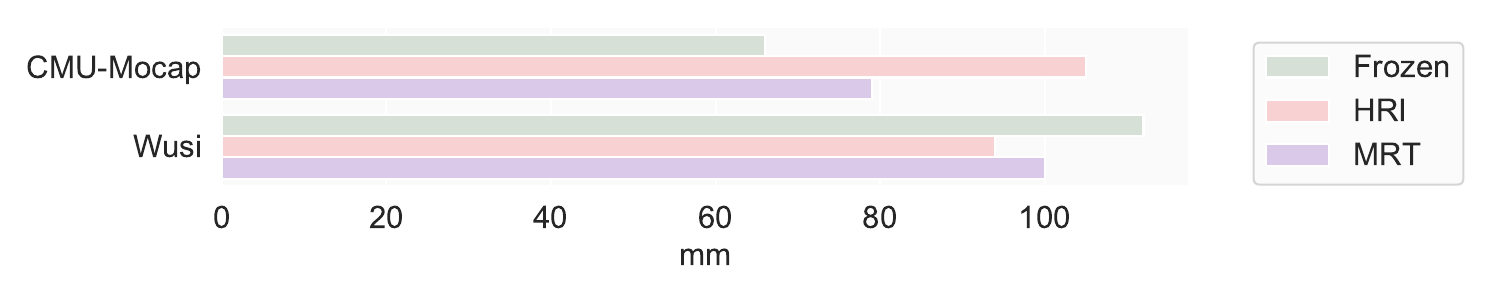}
  \caption{Comparison of baseline methods on CMU-Mocap~\cite{mocap} and our dataset. We compute the $1$s mean local pose prediction errors. The baseline methods are trained and tested on two datasets separately.} 
  \label{fig:baseline_bar}
\end{figure*}

\subsubsection{Quantitative results}

We first benchmark three baseline methods to assess the motion prediction difficulty of our dataset in comparison to the commonly-used existing dataset~\cite{mocap}. Figure ~\ref{fig:baseline_bar} reveals that for the CMU-Mocap dataset~\cite{mocap}, merely repeating the last-frame pose (\emph{Frozen}) leads to significantly lower errors than the model predictions. This finding indicates that the motions in this dataset are relatively ``static'', making it less challenging and informative for the social motion prediction task. In contrast, the dynamic nature of our dataset gives rise to considerably higher prediction errors of the naive \emph{Frozen} baseline. The results demonstrate that the proposed dataset serves as a more convincing benchmark for social motion prediction and introduces more challenges to the field.

Furthermore, we conduct a quantitative evaluation of the proposed framework's performance and compare it with state-of-the-art approaches. As illustrated in Table \ref{tab:methods}, our method achieves competitive results with SOTA approaches. It is worth noting that the single-person method HRI \cite{hri} demonstrates respectable short-term motion prediction accuracy; however, its performance over longer horizons is constrained by the absence of global context, particularly in root trajectory prediction. In contrast, our approach significantly surpasses SOTA methods in long-term social motion prediction, highlighting its effectiveness in modeling long-range human interactions. In the subsequent section, we will demonstrate the critical role that social context plays in predicting long-term human motion via examples.

\begin{table}[t]
\centering
\caption{Performance comparison with the baseline methods. We compute the mean prediction errors of global pose, local pose, and root, respectively.}
\label{tab:methods}
\resizebox{\linewidth}{!}{%
\scriptsize
\setlength{\tabcolsep}{4pt}
\begin{tabular}{l|cccc|cccc|cccc}
\toprule
milliseconds & \multicolumn{4}{c}{Global} & \multicolumn{4}{c}{Local} & \multicolumn{4}{c}{Root} \\
   & 400 & 600 & 800 & 1000 & 400 & 600 & 800 & 1000 & 400 & 600 & 800 & 1000 \\ \midrule
 Frozen & 119.2 & 165.7 & 207.3 & 244.5 & 62.8 & 82.8 & 98.9 & 112.1 & 99.8 & 139.3 & 175.0 & 207.6 \\
Social-STGCNN~\cite{mohamed2020social} & 188.2 & 246.4 & 298.8 & 346.5 & 62.8 & 82.8 & 98.9 & 112.1 & 173.1 & 225.6 & 273.7 & 317.9 \\
SocialVAE~\cite{socialvae2022} & 84.0 & 127.2 & 171.7 & 215.1 & 62.8 & 82.8 & 98.9 & 112.1 & 52.6 & 89.7 & 130.3 & 170.8 \\
 HRI~\cite{hri} & \textbf{50.0} & 106.8 & 145.5 & 189.2 & \textbf{39.5} & 62.9 & 78.4 & 94.1 & \textbf{40.1} & 88.0 & 119.6 & 157.7 \\
 MRT~\cite{mrt} & 66.9 & 103.2 & 140.2 & 176.4 & 49.4 & 68.8 & 85.4 & 99.5 & 50.6 & 79.0 & 109.3 & 140.1\\ 
 SoMoFormer~\cite{vendrow-somoformer} & 53.2 & 88.8 & 124.9 & 160.0 & 42.3 & 62.6 & 79.7 & 93.9 & 42.5 & 70.7 & 100.8 & 131.2 \\
Ours & 54.6 & \textbf{86.2} & \textbf{119.3} & \textbf{152.5} & 43.7 & \textbf{60.8} & \textbf{74.6} & \textbf{86.6} & 41.7 & \textbf{66.9} & \textbf{94.8} & \textbf{124.0} \\
 \bottomrule
\end{tabular}
}
\vspace{-0.1cm}
\end{table}

\begin{figure*}[t]
  \centering
  \includegraphics[width=\linewidth]{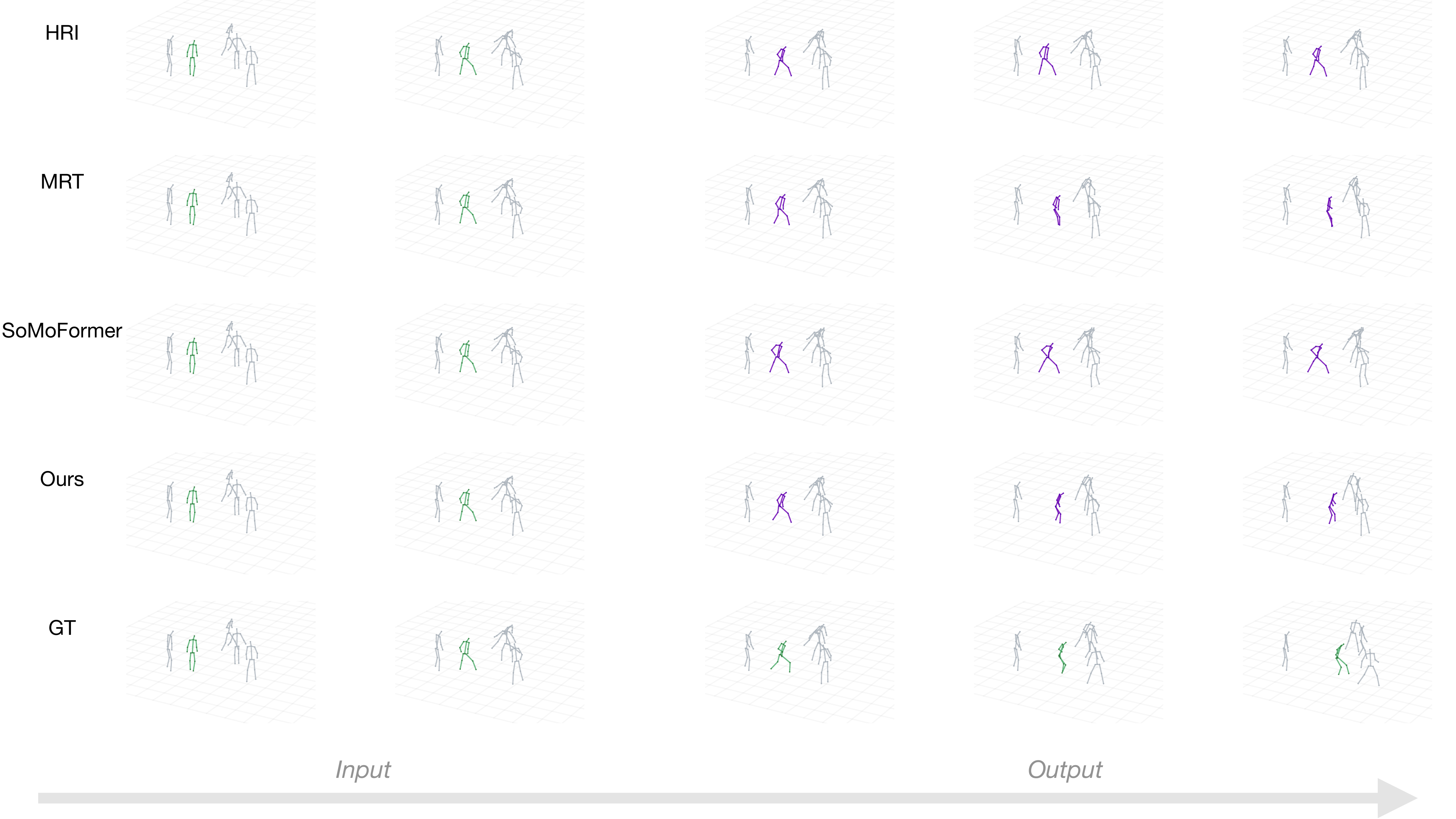}
  \caption{Qualitative comparison with existing methods. Left two columns are past motion and right three columns are future motion. We highlight the prominent player with bright colors. Real motion are shown in \textcolor[HTML]{6ebc85}{green}, and model predictions are shown in \textcolor[HTML]{8e39c6}{purple}.}
  \label{fig:sota}
  \vspace{-0.4cm}
\end{figure*}

\begin{figure*}[t]
  \centering
  \includegraphics[width=\linewidth]{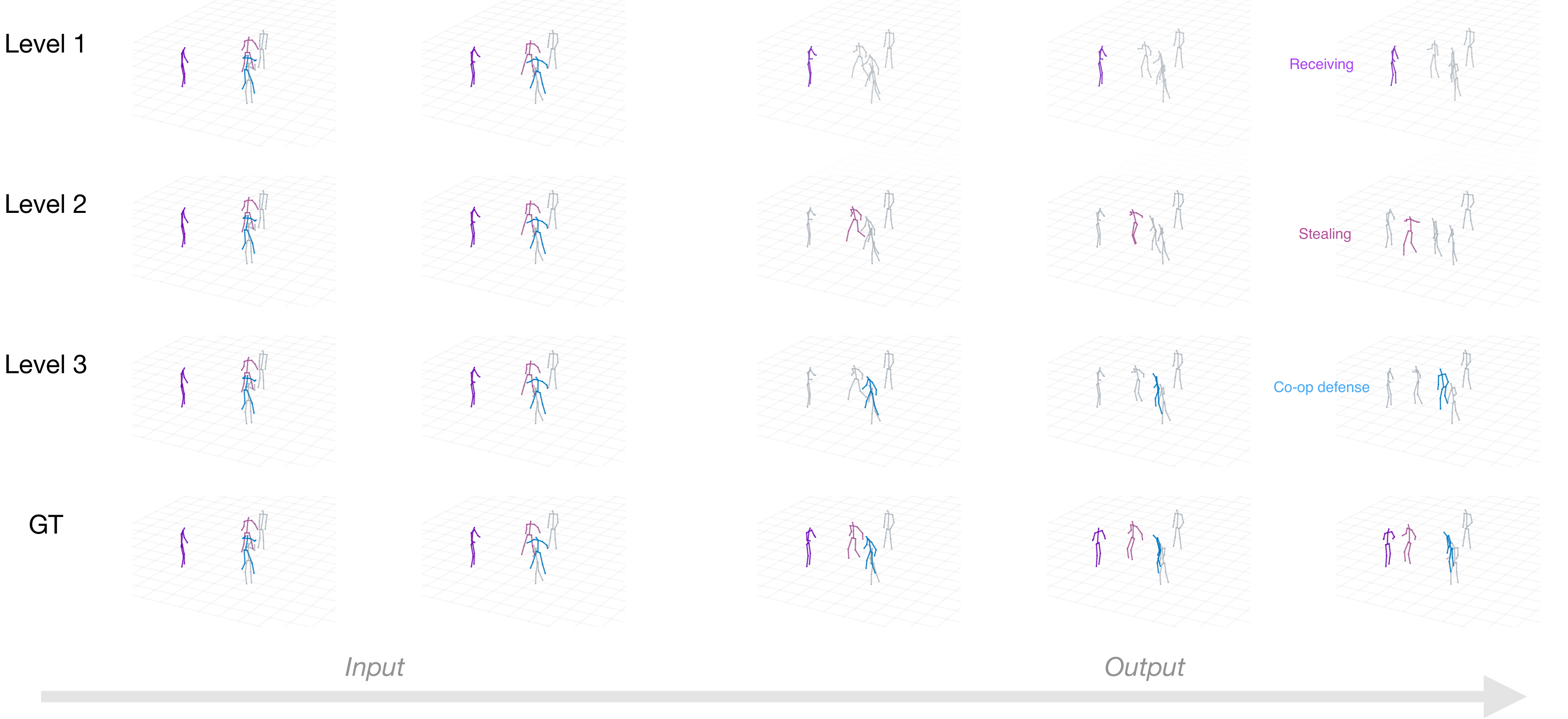}
  \caption{Actions produced by policy networks at different levels. Left two columns are past motions and right three columns are future motions. }
    
  \label{fig:levels}
  \vspace{-0.4cm}
\end{figure*}

\subsubsection{Qualitative results}

In order to better understand the performance improvement of our method, we visualize its motion prediction results and compare with the baseline methods qualitatively. Figure~\ref{fig:sota} illustrates an representative result in which the input motion shows that a defensive player (highlighted) is switching the target of defense. Our approach correctly predicts its future direction and orientations as the offensive player at the other side will likely be covered by the teammate. In contrast, other approaches fail to reason this, predicting that the player would continue to move along the input trajectory or simply stop at the middle.

In Figure~\ref{fig:levels}, three players to be noticed are highlighted in \textcolor[HTML]{b471a6}{rose}, \textcolor[HTML]{8e39c6}{purple} and \textcolor[HTML]{278acb}{blue} (we also call them this way below). 
Additionally, we visualize the actions produced by the intermediate policy networks to interpret the decision-making process. Shown on the top row, player \textit{\textcolor[HTML]{8e39c6}{purple}} at level $1$ policy network simply follows the past trend to receive the ball. As displayed in the second row, defender \textit{\textcolor[HTML]{b471a6}{rose}} at level $2$ gains knowledge from the first level that player \textit{\textcolor[HTML]{8e39c6}{purple}} is receiving the ball, and implements a steal attempt. Nevertheless, such aggressive actions could create defensive gaps. Defender \textit{\textcolor[HTML]{278acb}{blue}} at level $3$ senses such gaps caused by defender \textit{\textcolor[HTML]{b471a6}{rose}} from level $2$, therefore executes cooperative defense upwards to fill the gap.

\subsection{Ablation study}

Finally, we conduct ablation studies to investigate the influence of each component. In Table~\ref{tab:ablation} (a)-(e), we change the policy network depth $K$. $K=0$ corresponds to producing the final action directly without any recursive reasoning. We find that introducing the recursive reasoning process ($K>0$) clearly lowers the prediction errors, proving the effectiveness of our framework design based on cognitive hierarchy. From (f) to (h), we keep $K=3$ and study the influence of training regularizations. (f) removes the GAIL regularization for intermediate levels $1 \dots K-1$, and (g) completely removes GAIL. Training without GAIL increases the long-term trajectory prediction errors, while slightly lowers the local pose prediction error. We also observe that models without GAIL tend to produce motions that appear less natural (e.g. limb twisting), leading to inferior visual effects. In (h), we use different policy network instances for levels $1 \dots K$, and it turns out that sharing the policy networks leads to better generalization. 

In addition to the qualitative visualization of actions generated by policy networks at different reasoning levels, we also provide a quantitative comparison of their distance from ground-truth. Table~\ref{tab:levels} demonstrates that policy networks at different levels learn different actions, and policy networks at higher reasoning levels produce actions with less prediction errors. We hypothesize that, through the proposed recursive reasoning process, policy networks at different levels display a range of cognitive types, with behavior spanning from relatively naive to substantially rational. Higher-level policies more closely approximate an equilibrium strategy, which is usually in alignment with expert demonstrations. We provide several example motion sequences in the video demo to further illustrate this observation.

\begin{table}[t]
\centering
\caption{Ablation study of cognition levels and architecture designs.}
\label{tab:ablation}
\resizebox{\linewidth}{!}{%
\scriptsize
\setlength{\tabcolsep}{4pt}
\begin{tabular}{l|cccc|cccc|cccc}
\toprule
milliseconds & \multicolumn{4}{c}{Global} & \multicolumn{4}{c}{Local} & \multicolumn{4}{c}{Root} \\
   & 400 & 600 & 800 & 1000 & 400 & 600 & 800 & 1000 & 400 & 600 & 800 & 1000 \\ \midrule

(a) $K=0$ & 60.1 & 94.3 & 129.2 & 163.7 & 47.4 & 66.2 & 82.0 & 95.3 & 45.7 & 72.7 & 101.8 & 131.8 \\
 (b) $K=1$ & 55.2 & 88.0 & 121.9 & 155.5 & 44.3 & 62.1 & 77.0 & 89.7 & 42.3 & 68.0 & 96.3 & 125.4 \\
 (c) $K=2$ & 54.8 & 87.2 & 120.7 & 154.2 & 43.5 & 60.6 & 75.0 & 87.3 & \textbf{41.7} & 67.6 & 95.9 & 125.3 \\
 (d) $K=4$ & \textbf{54.5} & 86.3 & 119.7 & 153.5 & 43.6 & 60.5 & 74.7 & 86.6 & 42.0 & 67.3 & 95.5 & 125.2 
 \\
  \midrule
(e) $K=3$, full & 54.6 & \textbf{86.2} & \textbf{119.3} & \textbf{152.5} & 43.7 & 60.8 & 74.6 & 86.6 & 41.7 & \textbf{66.9} & \textbf{94.8} & \textbf{124.0} \\
\midrule
(f) \emph{w/o} mid GAIL & 54.9 & 87.5 & 121.4 & 155.4 & 43.7 & \textbf{60.4} & \textbf{74.5} & \textbf{86.4} & 41.8 & 67.8 & 96.6 & 126.9 \\
(g) \emph{w/o} GAIL & 54.6 & 87.1 & 121.1 & 155.1 & \textbf{43.4} & 60.4 & 74.7 & 86.8 & 41.9 & 68.1 & 97.0 & 127.0 \\
(h) \emph{w/o} weight sharing & 57.1 & 91.0 & 125.8 & 160.2 & 44.4 & 62.2 & 77.1 & 89.7 & 43.6 & 70.3 & 99.5 & 129.5 \\
 \bottomrule
\end{tabular}
}
\vspace{-0.4cm}
\end{table}

\begin{table}[t]
\centering
\caption{Quantitative comparison of actions produced by policy networks at different levels.}
\label{tab:levels}
\resizebox{\linewidth}{!}{%
\scriptsize
\setlength{\tabcolsep}{4pt}
\begin{tabular}{l|cccc|cccc|cccc}
\toprule
milliseconds & \multicolumn{4}{c}{Global} & \multicolumn{4}{c}{Local} & \multicolumn{4}{c}{Root} \\
   & 400 & 600 & 800 & 1000 & 400 & 600 & 800 & 1000 & 400 & 600 & 800 & 1000 \\ \midrule
Level-$1$ & 76.8 & 122.4 & 170.3 & 217.7 & 50.1 & 69.0 & 85.2 & 99.1 & 61.4 & 100.4 & 143.2 & 186.9 \\
Level-$2$ & 61.0 & 99.9 & 144.6 & 192.5 & 47.1 & 66.0 & 82.5 & 96.6 & 46.5 & 78.0 & 116.2 & 158.8 \\
Level-$3$ & \textbf{54.6} & \textbf{86.2} & \textbf{119.3} & \textbf{152.5} & \textbf{43.7} & \textbf{60.8} & \textbf{74.6} & \textbf{86.6} & \textbf{41.7} & \textbf{66.9} & \textbf{94.8} & \textbf{124.0} \\
 \bottomrule
\end{tabular}
}
\end{table}

\section{Conclusion}
\label{sec:conclusion}

In this work, we propose the first large-scale multi-person 3D motion dataset featuring strategic human social interactions. This dataset surpasses the existing ones in scale, diversity, dynamics, and interaction, thus posing new challenges to the social motion prediction problem. We reformulate the problem using a MARL perspective. Building on this, we propose a framework that effectively combines behavioral cloning (BC), generative adversarial imitation learning (GAIL), and cognitive hierarchy. Our approach demonstrates strong generalization capabilities and improved interpretability for modeling strategic human social interactions. 

\textbf{Limitations:} Our approach is based on the cognitive hierachy assumption and may not hold true for all type of social interactions. This work do not explicitly model social roles, which can be explored as future work.

\paragraph{Acknowledgment} This work was partially supported by National Key R\&D Program of China (2022ZD0114900). We thank Xuesong Fan and Rujie Wu for their generous support during the data collection process.

\clearpage

{\small
\bibliographystyle{ieee_fullname}
\bibliography{ref}
}

\end{document}